# Fitness Dependent Optimizer with Neural Networks for COVID-19 patients

## Abstract

The Coronavirus, known as COVID-19, which appeared in 2019 in China, has significantly affected global health and become a huge burden on health institutions all over the world. These effects are continuing today. One strategy for limiting the virus's transmission is to have an early diagnosis of suspected cases and take appropriate measures before the disease spreads further. This work aims to diagnose and show the probability of getting infected by the disease according to textual clinical data. In this work, we used five machine learning techniques (GWO_MLP, GWO_CMLP, MGWO_MLP, FDO_MLP, FDO_CMLP) all of which aim to classify Covid-19 patients into two categories (Positive and Negative). Experiments showed promising results for all used models. The applied methods showed very similar performance, typically in terms of accuracy. However, in each tested dataset, FDO_MLP and FDO_CMLP produced the best results with 100% accuracy. The other models' results varied from one experiment to the other. It is concluded that the models on which the FDO algorithm was used as a learning algorithm had the possibility of obtaining higher accuracy. However, it is found that FDO has the longest runtime compared to the other algorithms. The link to the covid 19 models is found here: https://github.com/Tarik4Rashid4/covid19models

## 1. Introduction

The emergence of the new Coronavirus at the end of 2019 in Wuhan/China led to a global crisis and posed a great challenge for the world to deal with it. The early reports predicted the outbreak of this virus based on its reproduction behaviour [1]. And as was expected, the various spread throughout China and its neighbouring countries in less than a month [2]. According to the WHO situation reports by the end of January 2020, there were more than 9800 confirmed cases and over 200 deaths [3]. The outbreak continued rapidly, reaching over a million confirmed cases, and more than 56000 deaths. At the beginning of April [4].

The challenges of dealing with this pandemic come for several reasons, including the ease of transmission from one person to another, as transmission occurs simply by direct contact with an infected person or through droplets resulting from sneezing and coughing [1]. What increases the seriousness of the situation is the possibility of transmission before the appearance of the symptoms, since the incubation period for the Corona ranges between 2 to 14 days, and the appearance of symptoms occurs approximately 12 days after infection, in addition to some cases that do not show any symptoms [5]. These reasons have led governments to impose preventive measures such as quarantine and social distancing. However, this procedure has psychological and economic damage to society [6]. ".

The motivation for this work mainly came from the impact of the virus on the world and the damage it has caused in health, social, and even economic terms. Recently, many contributions have appeared to employ techniques such as Artificial Intelligence (AI) to deal with this pandemic. One of the most common fields that have been focused on was using AI to build classifiers for classifying patients if they have COVID-19 or how severe their infection is. Classifiers are simply defined as grouping objects into classes or categories. The main purpose of classification is to have a predictive model which can identify the class or the category of the new data that falls under it. The Artificial Neural Network (ANN) is one of the most common classification methods and it has excelled a lot in research related to the new pandemic. Neural networks fall under the category of supervised machine learning, which means they require a training process where the network can identify patterns by providing data samples with inputs and known outputs so that they can generalize solutions, and thus the network can provide outputs close to the expected outputs for any given set of input values. This process of training is done by employing techniques to adjust the weights and thresholds of the neurons to generalize the solutions produced by their outputs [7].

For a long time, the majority of the studies that apply artificial neural networks use gradient optimization techniques to train the network, typically the Backpropagation neural network. The Backpropagation method has been applied to different applications, such as image processing, function approximation, and pattern recognition. However, this method has some serious drawbacks, most notably its slow coverage and ease of getting trapped in local solutions [8]. To overcome these problems, several techniques have been developed. The most common solution was to employ meta-heuristic algorithms [9]. Metaheuristic optimization has attracted many researchers because of its features that give them superiority over classical algorithms. One of these features is avoiding the local optimum trap. Metaheuristics are also known for solving multiple objective and nonlinear problems. Many algorithms have been developed in the past two decades, which almost gives unlimited options to come up with new techniques that serve the purpose of classification. There is already a huge number of available techniques to develop a classification model. However, the large diversity of these techniques makes it difficult to select the most efficient one. In addition, there are multiple evaluation criteria which makes it more challenging to decide whether one technique is better than the other [10].

The main contribution of this work is providing three data sets. Each data set will be tested using five intelligent models to perform the automatic classification of COVID-19-infected patients. Two types of neural networks will be tested with three nature-inspired algorithms, which are Fitness Depended Optimization (FDO), Grey Wolf Optimization (GWO), and Modified Grey Wolf Optimization (MGWO). The applied models aim to predict positive and negative Covid-19 cases. The objective of this work is to provide a solution through a system that can make a classification of the diagnosis through AI techniques and algorithms.

The rest of this paper is organized as follows: In section 2, we discuss the previous literature and the proposed methods that deal with COVID-19. Then in section 3, we discuss the preliminaries and the details of each method used in this work, and in section 4, the methodology of creating this work will be explained and how the proposed methods are applied. Section 5 will present the obtained results. Finally, we give our conclusion in section 6.

## 2. Related Work

With the COVID-19 pandemic continuing to spread, many experts from around the world are attempting to learn more about this novel disease and understand its behavior, ways to suppress its spread, and the appropriate treatment. In this section, we highlight the research that harnessed artificial intelligence to find solutions related to the virus, focusing more on the classification to predict the COVID-19 positive and negative cases, as this is the aim of our work. Based on the data type of the reviewed research works, this section will be split into three sections. Section 2.1 describes the literature that proposed the application of analyzing medical image data. Section 2.2 describes the literature that proposed the application of analyzing COVID-19 text data. Section 2.3 describes other applications that were used to deal with COVID-19.

### 2.1. Artificial Intelligence Applications in Medical Images

Most of the literature that discusses AI applications with COVID-19 deals with medical images, exclusively CT scans and X-ray images. The rapid outbreak of Covid has led to the urgent need for diagnosis and appropriate treatment of patients. For this, X-rays and CT scans are commonly used. In general, medical images take time for specialists to understand and give proper diagnoses, especially chest CT scans, which contain multiple slices. Thus, AI-based diagnosis with medical image classification is highly popular [11] . One of these works of literature is made by [12] who proposed a deep learning technique to analyze images of chest radiographs of COVID-19 patients treated in China and the United States. The deep learning method was implemented as a U-Net trained with 22000 radiographs that produce pneumonia probability color maps. The suggested technique's goal is to be useful in early diagnosis, as well as longitudinal and long-term monitoring of suspected pneumonia patients, including those with COVID-19 pneumonia.

[13] proposed CoroNet CNN, which is based on Xception architecture pre-trained on the ImageNet dataset to detect COVID-19. They performed two types of multi-classification. first with four classes (Normal, Covid, Pneumonia bacterial, and pneumonia viral). The second is with three classes (Normal, Covid, and Pneumonia). The overall model accuracy was 89.6%.

[14] proposed a convolutional neural network that is based on the concatenation of Xception and ResNet50V2. Their model performs multi-classification of X-ray images categorized into three classes: normal, Covid and pneumonia. The overall model accuracy was 91.4%.

Some research works proposed methods to make real-time implementation systems, such as the techniques proposed by [15] and [16]. Both applied Convolutional Neutral Network (CNN) and Extreme Learning Machine (ELM) to detect COVID-19 from X-ray images in real time. The difference between their methods is the metaheuristic algorithms that have been used to stabilize the ELM. [15] suggested the Sine-Cosine Algorithm optimization algorithm to build their CNN model, which achieved an accuracy of 98.83%. On the other hand, [16] used the Chimp Optimization Algorithm (ChOA) in their model. The accuracy that was achieved using the same database was 98.25%.

Many other methods were presented by different researchers, all performing classification tasks to detect COVID-19. Some of these proposals are summarized in Table 1. Although most of these models achieve promising results, they still need clinical study and testing.

*Table 1: Proposed models for medical image classification.*

| Ref. | Method and Application Summary | Result |
|---|---|---|
| [17] | Develop a hybrid deep neural network (HDNN), using computed tomography (CT) and X-ray imaging, to perform multi-class classification | The accuracy obtained is 99% |
| [18] | Presenting the CoVIRNet method (COVID Inception-ResNet model), which utilizes the chest X-rays to identify COVID-19 patients by performing multiclass classification (COVID-19, Normal, Pneumonia bacterial, and Pneumonia viral) | The best accuracy obtained is 97.29%. |
| [19] | Using Convolutional capsule networks (CapsNet) to perform binary classification (normal, and Covid), and multi-class classification (normal, Covid and pneumonia) to detect COVID-19 from chest X-ray images | An accuracy of 97.24% for binary classification. An accuracy of 84.22% for multi-class classification. |
| [20] | Presenting OptCoNet method, which is CNN and Grey Wolf Optimizer for hyperparameters optimization in training the CNN layers to perform Multi-class classification (normal, Covid, and pneumonia) to detect COVID-19 from chest X-ray images. | The best accuracy obtained is 97.78%. |
| [21] | Using a convolution network with a Bayesian algorithm for training and optimization to perform binary classification (normal, and Covid), and multi-class classification (normal, Covid and pneumonia) to detect COVID-19 from chest X-ray images. | An accuracy of 100% for binary classification. An accuracy of 98.3% % for multi-class classification. |
| [22] | DarkNet model which is a convolution network to perform binary classification (normal, and Covid), and multi-class classification (normal, Covid and pneumonia) to detect COVID-19 from chest X-ray images. | An accuracy of 98.08 % for binary classification. An accuracy of 87.02 % for multi-class classification |
| [23] | Applying Dense Convolutional Networks and transfer learning to perform Multi-class classification (normal, Covid, and pneumonia) to detect COVID-19 from chest X-ray images. | The best accuracy obtained is 100%. |
| [24] | Transfer learning with CNNs Multi-class classification (normal, Covid, and pneumonia) to detect COVID-19 from chest X-ray images. | The best accuracy obtained is 96.78%. |
| [25] | Applying for Transfer, and Compose (DeTraC) convolutional neural network to perform multi-class classification (norm 1, norm 2, COVID-19 1, COVID-19 2, SARS 1, and SARS 2) from chest X-ray images | The best accuracy obtained is 93.1% |
| [26] | Applying residual neural network with a total of 50 (ResNet50) convolutional neural networks to perform multi-class classification (Normal, Covid, Pneumonia bacterial, and pneumonia viral) from chest X-ray images | The accuracy obtained is 96.23% |
| [27] | Comparing three convolutional neural networks based (ResNet50, InceptionV3, and Inception- ResNetV2) by applying them to perform binary classification (normal, and Covid) from chest X-ray images | The accuracy of the ResNet50 model is 98% The accuracy of InceptionV3 is 97% The accuracy Inception-ResNetV2 is 87% |
| [28] | Developing a 3D deep learning framework called COVnNet which consists of a RestNet50 to perform multi-class | The sensitivity for detecting COVID-19 is 90% |
| classification (Normal, Covid, Pneumonia (CAP)) from chest CT | Community-Acquired | The specificity for detecting COVID-19 is 96% |
| | | The sensitivity for detecting CAP is 94% |
| | | The specificity for detecting CAP is 96% |

## 2.2 Artificial Intelligence Applications in Patient Records

Diagnostic research usually uses medical imaging, and there are very few studies that diagnose using patient records compared to image data. Some approaches don't require imaging equipment for COVID-19 tracking and diagnosing. One of these approaches was by [29]. They compared different machine learning techniques (Decision Tree Extremely Randomized Trees, K-nearest neighbors, Logistic Regression, – Naive Bayes, Random Forest) in classifying patients into two categories (Normal, Covid) based on information from the patient's blood tests. They also developed a modified random forest method called a three-way Random Forest classifier which reported higher performance in terms of accuracy of 86%.

Another work was presented by [30]. They classified clinical reports that contained information about the symptoms of COVID-19 and other viruses. They classified their data into four classes: Covid, SARS, ARDS, and both (SARS and COVID) using classical machine learning techniques (Logistic regression, Multinomial Naive Bayesian, Support vector machine, and Decision tree). The highest accuracy obtained was by the Logistic regression and Multinomial Naive Bayes models with 96.2% testing accuracy. Then the decision tree comes in second place with 92.5% testing accuracy. Finally, support vector machine with 90.6% accuracy.

[31] also applied traditional machine learning methods including (logistic regression, decision tree, support vector machine, naive Bayes, and artificial neural network) to classify patients into positive and negative COVID-19 infections based on information about chronic diseases the patient has. The result of the models applied showed that showed the decision tree model has the highest accuracy of 94.99% followed by Logistic regression with 94.41% then naive Bayes with 94.36%, support vector machine with 92.4, and finally artificial neural network with 89.2%

In the last three discussed research papers, there is a discrepancy in the results even though the same models are used in each work. This discrepancy is expected because each work used a different dataset and a different number of classes.

We observed that papers that discuss diagnosing with datasets that contain patient records use only traditional machine learning techniques, and there was no attempt to develop more effective and sophisticated models like the ones that are presented in image classification. In our work, we aim to build a reliable and efficient model to predict the negative and positive COVID-19 based on the patient records dataset.

## 2.3 Other Artificial Intelligence Applications

Many interesting solutions have been proposed that do not rely solely on diagnosing the disease based on RT-PCR data and chest X-ray tests. A solution was presented by [32] to detect the disease by analyzing the audio data using the Gradient Boosting Machine-based classifier. The methods that have been used in this research were the LGM classifier, Random Forest (RF), SVM, and K-Nearest Neighbor (KNN). The obtained results showed the overall average accuracy was above 97%.

Detecting protective measures such as masks, goggles, and protective clothing is also a very important step in the fight against COVID-19. [33] proposed a very interesting solution that uses unmanned vehicles (UV) to build a map of the real environment and detect protective measures, body temperature measurements, and other advanced tasks.

# 3. Preliminaries

In this research Grey wolf Optimizer, Modified Grey wolf Optimizer and Fitness Dependent Optimizer have been employed to find the optimum weight and bias for a neural network to predict the covid-19 negative and positive patients. The details of each algorithm are explained in this section.

## 3.1 Grey Wolf Optimization

Grey Wolf Optimization is a swarm intelligence algorithm and one of the most powerful algorithms that were proposed by Mirjalili et al. in 2014 [34]. The GWO is one of the nature-inspired algorithms, which was inspired by grey wolves' behaviour during hunting. In nature, Grey wolves seek out the most efficient approach to hunting prey using a certain procedure. The GWO algorithm uses the same process that is used by the grey wolves in nature to organize the diverse responsibilities in the wolves' pack, which follows the pack hierarchy. The social hierarchy is interpreted as the optimality of solutions. The most preferred solution is the alpha ($\alpha$) Consequently, the second and third-best solutions are beta ($\beta$) and delta ($\delta$) followed by the other solutions categorized as omega ($\omega$).

During the hunt, the grey wolf will encircle the prey first. Equation (1) models the updating of the grey wolf position ($\vec{X}(t)$) around the prey ($\vec{X}_p(t)$) [34]:

$$\vec{D} = |\vec{C}.\vec{X}_p(t) - \vec{X}(t)| \tag{1}$$

$$\vec{x}(t+1) = \vec{X}_p(t) - \vec{A}.\vec{D} \tag{2}$$

Where t is the current iteration. $\vec{A}$ and $\vec{C}$ are the coefficient Vectors. $X_p$ is the position vector of the prey and $X$ is the position vector of the grey wolf.

The $\vec{A}$ and $\vec{C}$ vectors are represented by the equations [34]:

$$\vec{A} = 2\vec{a}.r_1 - \vec{a} \tag{3}$$

$$\vec{C} = 2.\vec{r}_2 \tag{4}$$

Where the components of $\vec{a}$ are linearly decreased from 2 to 0 throughout iterations and $r_1, \vec{r}_2$ are random vectors in [0, 1].

with equations (3) and (4) the Grey wolf/ search agent can update its position in any location. by adjusting the value of $\vec{A}$ and $\vec{C}$ vectors the best grey wolf can be reached as shown in Figure 1 where the wolf in the position of $(X, Y)$ can relocate itself around the prey $(X^*, Y^*)$ with the proposed equations.

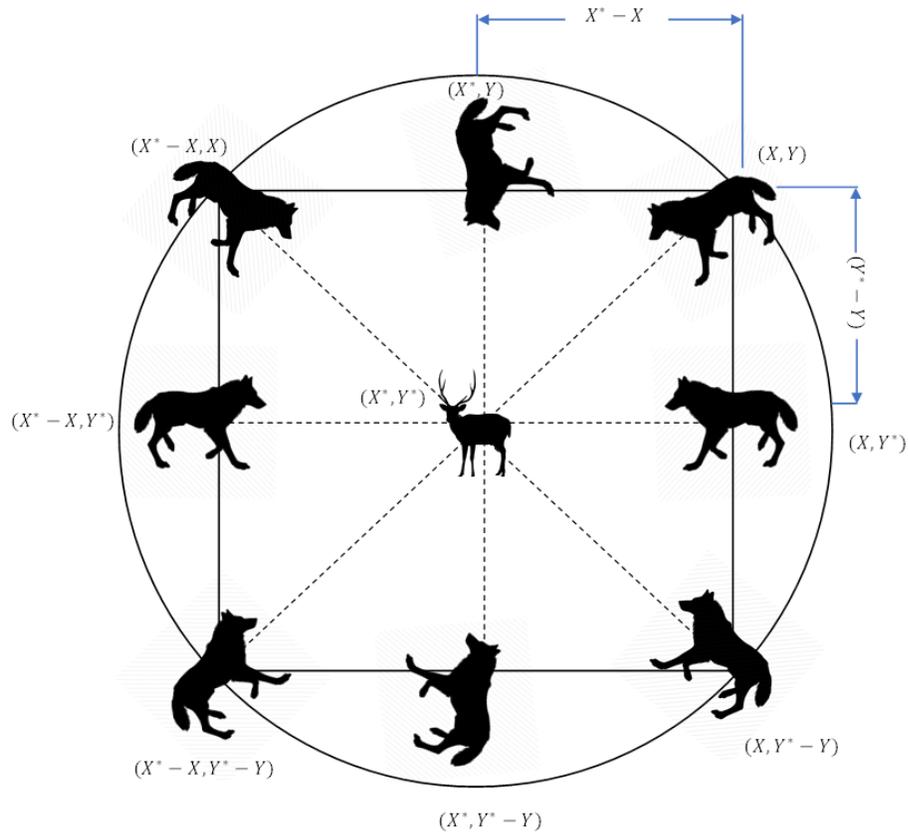

*Figure 1: Position vectors and their possible next locations in 2d space.*

In nature, grey wolves can recognize the location of their prey, but when it comes to mathematical application, the location of the prey (the optimum solution) is unknown. In this case, we assume that the alpha, beta, and delta have better knowledge about the potential location of prey. Therefore, the solutions that are obtained by these three search agents will be saved and the other agents (omegas) will update their positions according to the position of the saved ones. This process is represented by the following equations where the distance between alpha, beta, and delta is calculated [34]:

$$\vec{D}_\alpha = |\vec{C}_1 \cdot \vec{X}_\alpha - \vec{X}| \tag{5}$$

$$\vec{D}_\beta = |\vec{C}_2 \cdot \vec{X}_\beta - \vec{X}| \tag{6}$$

$$\vec{D}_\delta = |\vec{C}_3.\vec{X}_\delta - \vec{X}| \tag{7}$$

Where $\vec{X}_\alpha, \vec{X}_\beta,$ and $\vec{X}_\delta$ are the positions of alpha, beta, and delta. $\vec{X}$ is the current solution's position [34].

$$\vec{X}_1 = \vec{X}_\alpha - \vec{A}_1.(\vec{D}_\alpha) \tag{8}$$

$$\vec{X}_2 = \vec{X}_\beta - \vec{A}_2.(\vec{D}_\beta) \tag{9}$$

$$\vec{X}_3 = \vec{X}_\delta - \vec{A}_3.(\vec{D}_\delta) \tag{10}$$

$$\vec{X}(t+1) = \frac{\vec{X}_1 + \vec{X}_2 + \vec{X}_3}{3} \tag{11}$$

The main controlling parameter of GWO to promote exploration is the C Vector. the value of this parameter is a random value in the interval of [0, 2]. Its purpose is to give the prey a random weight depending on the position of a wolf. this parameter also makes the process of reaching the prey harder and farther ($C > 1$) or easier and closer ($C < 1$). This component is very helpful in avoiding local optima stagnation, especially in the final iterations.

Another controlling parameter that causes exploration is $A$. The value of $A$ parameter is defined based on the value $a$ parameter, which linearly decreases from 2 to 0. If the random values of $\vec{A}$ are within the $[-1.1]$ interval, then the next position of a wove can be in any position between its current position and the prey position. Thus when $|\vec{A}| > 1$ the wolves attack the prey and when $|\vec{A}| < 1$ the wolves attack far from the prey (see Figure 2).

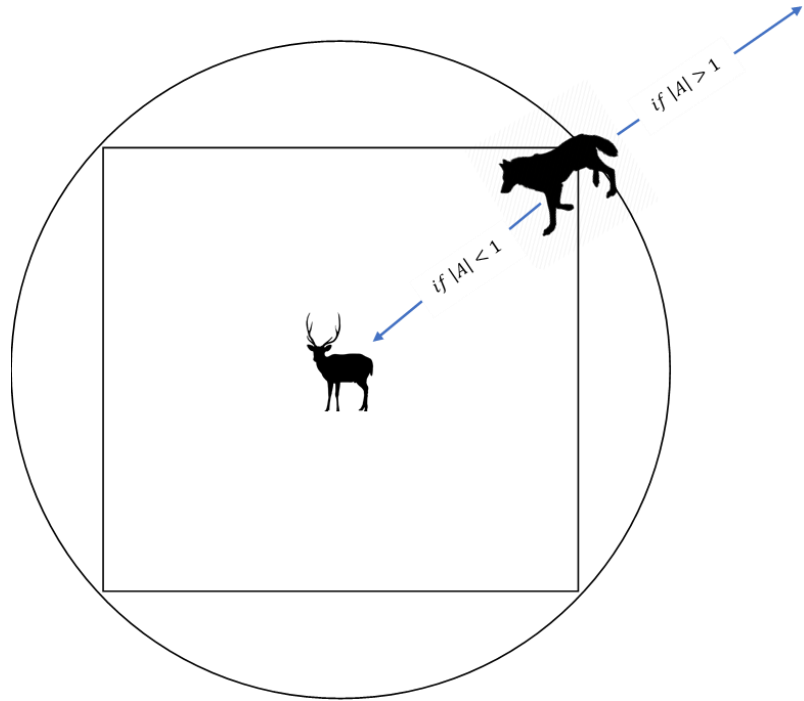

*Figure 2: Attacking prey versus searching for prey.*

## 3.2 Modified Grey Wolf Optimization

In 2019 [35] proposed a modified version of Grey Wolf Optimizer. The proposed algorithm has two simple modifications. The first modification was instead of having four groups of wolves Alpha ($\alpha$), Beta ($\beta$), Delta ($\delta$), and Omega ($\omega$). Extra group called Gama ($\gamma$) were added. With this extra group, the omega wolved will update their position concerning the position of four wolves (Alpha, Beta, Delta, and Gamma) instead of three.

The second modification is regarding the position defining the step size of the omega wolves which is shown in standard GWO in equations (8), (9), (10), and (11). In this modified version another equation will be added to calculate the distance between alfa, beta, delta, and gamma [35] :

$$\vec{D}_\gamma = \left| \vec{C}_4 . \vec{X}_\gamma - \vec{X} \right| \qquad (12)$$

Where $\vec{X}_\gamma$ : The position of gamma. $\vec{X}$ : The current solution.

Then the positions of alfa, beta, delta, and gamma will be calculated as the following [35]:

$$\vec{D}_{avg} = \frac{\vec{D}_1 + \vec{D}_2 + \vec{D}_3 + \vec{D}_4}{4} \tag{13}$$

$$\vec{X}_1 = \vec{X}_\alpha - \vec{A}_1 \cdot (\vec{D}_{avg}) \tag{14}$$

$$\vec{X}_2 = \vec{X}_\beta - \vec{A}_2 \cdot (\vec{D}_{avg}) \tag{15}$$

$$\vec{X}_3 = \vec{X}_\delta - \vec{A}_3 \cdot (\vec{D}_{avg}) \tag{16}$$

$$\vec{X}_4 = \vec{X}_\gamma - \vec{A}_4 \cdot (\vec{D}_{avg}) \tag{17}$$

And finally, the equation that represents the current solution's final position is [35]:

$$\vec{X}(t+1) = \frac{\vec{X}_1 + \vec{X}_2 + \vec{X}_3 + \vec{X}_4}{4} \tag{18}$$

## 3.3 Fitness Dependent Optimizer

Fitness Dependent Optimizer is one of the most recent natural-inspired algorithms that has been proposed. It was developed by Jaza Abdullah and Tarik Rashid in 2019. The algorithm was inspired by the swarm of bees' behaviour during reproduction. The FDO algorithm mimics the process of scouting bees when searching for a suitable hive among many potential hives to find suitable solutions among possible solutions [36]. Compared to GWO, the FDO has a simpler concept and is easier to understand. The FDO process can be divided into two parts: the searching process, where the search agents attempt to find the best solution, and the movement process, where the scout bee updates its position [37]. These two processes will be explained in detail in the coming sections.

### 3.3.1 Scout bee searching process

The main essence of this process is to find suitable new hives "solutions". As with the GWO algorithm, this algorithm uses search agents to search for new hives known as scout bees. Finding a new solution in this algorithm is represented by a scout bee position. At the beginning of the FDO execution, the locations of the artificial scout bees are initiated randomly in the search spaces. Then, throughout the execution, the global best solution is determined by the algorithm. The scout bees use a combination of a random walk and a fitness weight mechanism. to search for new hives/solutions in the search space. The scout bees will keep searching for better solutions till the end of the determined boundaries. If a better solution is found, the previous one is simply ignored. And if no better solution is found, it will keep the former solution.

### 3.3.2 Scout bee movement process

In FDO, the scout bees update their position to obtain a better solution. The scout bees' position is updated by adding pace to the current position as shown in equation (19) [36]:

$$X_{i,t+1} = X_{i,t} + pace \tag{19}$$

Where $i$ is The current search agent. $t$ is the current iteration. $X$ is the artificial scout bee (search agent) and $pace$ id the movement rate and direction of the artificial scout bee.

The pace relatively depends on the fitness weight ($fw$). Nevertheless, the direction of pace is completely dependent on a random mechanism. The determination of ($fw$) is done according to equation 20 [36]:

$$fw = \left|\frac{x^*_{i,tfitness}}{x_{i,tfitness}}\right| - wf \tag{20}$$

Where $x^*_{i,tfitness}$ represent the best global solution's fitness function value. $x_{i,tfitness}$ is the current solution's value of the fitness function and $wf$ is the weight factor.

The weight factor ($wf$) is used to control $fw$ and its value is either 0 or 1. If $wf$ is equal to 1, then it represents a high level of convergence and a low chance of coverage. and if $wf$ equals 0, then it will be neglected because it will not affect the equation (20). Setting the value of $wf$ to 0 doesn't necessarily make the search more stable. In some cases, the opposite occurs as the fitness function value depends on the problem.

the FDO has to consider some settings for ($fw$) to avoid unacceptable cases such as making sure that the $fw$ value is in the [0, 1] range. as well as avoiding the division by zero which can occur if the value of $x_{i,tfitness}$ is 0. Therefore, the following rules which are represented in the equations

(21) should be used [36]:

$$\begin{cases} fw = 1 \text{ or } fw = 0 \text{ or } x_{i,t \text{ fitness}} = 0, pace = x_{i,t} * r \\ fw > \text{ and } fw < 1 \begin{cases} r < 0, pace = (x_{i,t} - x_{i,t}^*) * fw * -1 \\ r \geq 0, pace = (x_{i,t} - x_{i,t}^*) * fw \end{cases} \end{cases} \quad (21)$$

Where:

$r$ is a random number in the range of [-1, 1]. $x_{i,t}$ is the current solution. $x_{i,t}^*$, is the global best solution achieved sofar.

## 4. Research Methodology

In this work, different methods have been applied to be compared. There are five models applied, each with a different network architecture or different training algorithms. Each of these models was tested on three datasets. The methodology of this work consists of defining the dataset, preparing the data (if it needs translation or representation), feature selection, applying the classification model, and making statistical comparisons (See Figure 3).

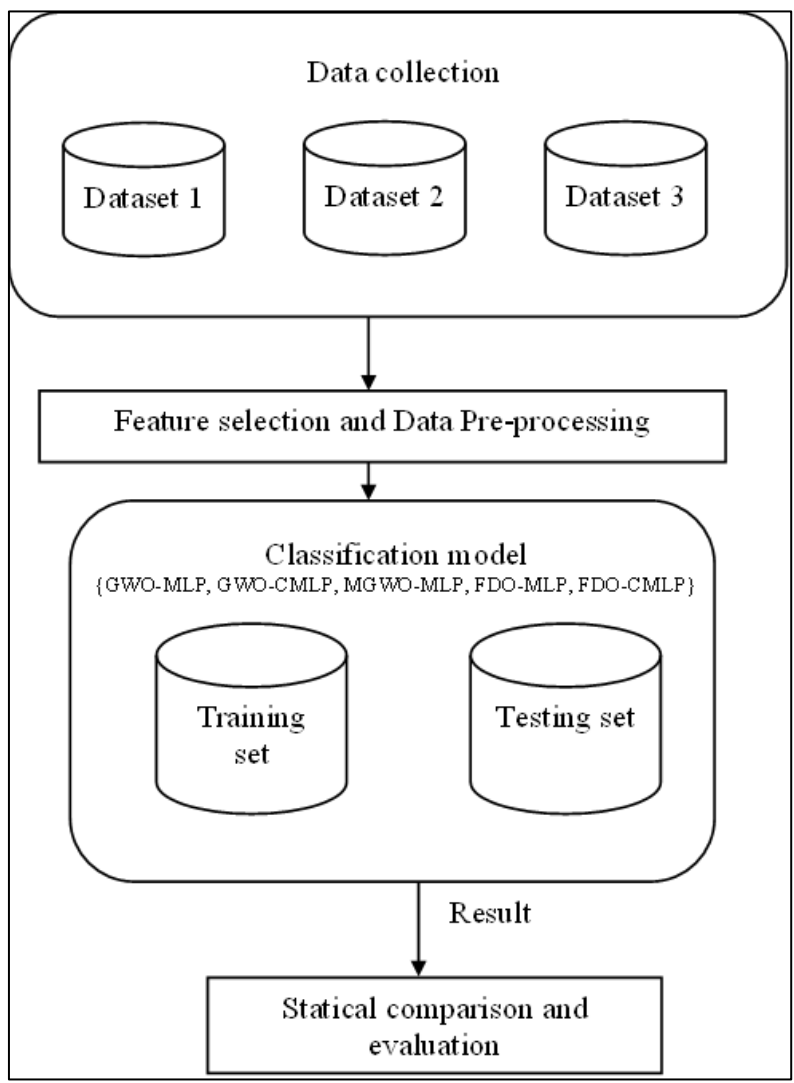

Figure 3: research methodology

## 4.1 Data collection

Building a database is the most important step in machine learning. Three different datasets have been collected in this work. The first dataset was collected by [38]. This data is an original Brazilian COVID-19 dataset that contains early-stage symptoms, comorbidities, and demographic information for patients tested in Brazil. The dataset was compiled with records from 26 Brazilian states and federations. Testing was done by viral and antibody tests.

The second dataset, which is available in [39], also provides early-stage symptoms and comorbidities based on various WHO guidelines, in addition to additional information about the preventive measures followed by the patient.

The third dataset contains demographic and clinical data as well as results of the RT-PCR test for COVID-19 in patients with a viral respiratory diagnosis in Mexico. Mexico as reported by the [General Directorate of Epidemiology]. The dataset is available in [40].

## 4.2 Feature Selection and Data Pre-processing

After data collection, the relevant features were selected. Some records were removed from the dataset either because they were duplicated or because they contained missing or ignored information. Another action that has been made is translating the dataset that contains records of patients from Mexico from Portuguese to English. Some of the datasets originally represented the positive and negative responses with "yes" for positive and "no" for negative. To prepare the data to be used in the models, we represented the positive input feature with 0 and the negative input feature with 1. As for the target, 1 represented "positive" and 2 represented "negative".

## 4.3 Classification Model

In this section, we describe the methodology of the proposed models. This includes the architecture used to build the models, which is covered in Section 4.3.1, and the training methods that are described in Section 4.3.2.

### 4.3.1 The architecture of the neural network

Three elements define the architecture of neural networks, which are the number of layers of processing elements or nodes, including input and output, the number of hidden layers, and the number of nodes in each layer. Determining the topology of a neural network lies in controlling the number of hidden layers and the neurons in each layer [41].

In all of the proposed models, we used a single hidden layer with the number of neurons determined based on the number of features that the dataset has (see Table 2). The number of neurons in the hidden layer is set by the following rule:

$$Hno = 2*Ino + 1 \tag{22}$$

Where Hno is the number of hidden layers and Ino is the number of input layers.

*Table 2 number of input and hidden neurons based on the dataset that is used in each model.*

| Dataset | Number of Input neurons (features) | Number Hidden Neurons |
|---------|-----------------------------------|----------------------|
| Dataset 1 | 10 | 21 |
| Dataset 2 | 18 | 37 |
| Dataset 3 | 13 | 27 |

In this work we have applied two types of neural networks the first one is basic feed-forward artificial neural networks which consist of three interconnection layers: one input layer, a hidden layer, and one output layer. The other type is a cascade feed-forward artificial neural network. In both types, the ANN connects from the input layer to each layer and from each layer to the successive layers. The difference is in cascade ANN has an extra connection from the input layer directly to the output layer (see Figure 4). The additional connections are able the network to learn associations of high complexity as well as improve the speed at which the network learns the desired relationship [42] [43].

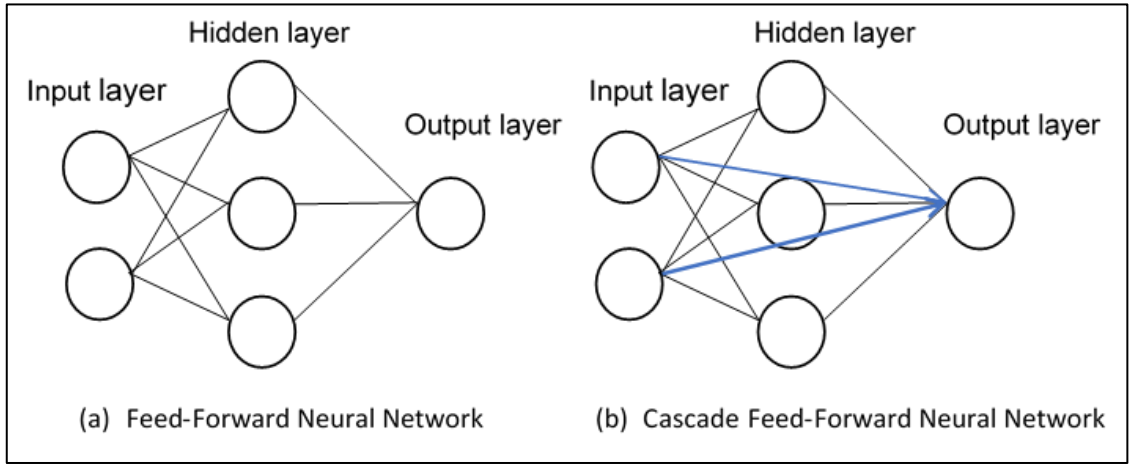

*Figure 4: Difference between feed-forward neural network and cascade feed-forward neural network.*

### 4.3.2 Artificial Neural Network Learning Method

The purpose of training is to find the set of values for weights and biases that provide the best classification accuracy. In this work, different models will each use one of the two metaheuristic algorithms (GWO, modified GWO, and FDO) as a training algorithm.

Since the best values for the weights and biases need to be found, they are defined as variables. After defining the variables for the optimization, we need an objective function for the algorithm. In this case, we will use one of the most common evaluation metrics for neural networks, which is Mean Square Error (MSE). The MSE gives the average or means of the square of the difference between the desired output and the value that is obtained from the ANN model. The following represents the calculation of the MSE:

$$MSE = \sum_{i=1}^{n}(y_i - \hat{y}_i)^2 \qquad (23)$$

Where n is the number of outputs. $i$ is the input unit iteration. $y_i$ represent the desirable output and $\hat{y}_i$ represents the obtained value.

To evaluate the performance of the model, we calculate the MSE over all the training samples, then take the average of the MSE. Based on the average value of MSE that is given to the optimization algorithm, the model will adapt itself and change the weights and biases to minimize the average MSE of all training samples. The average MSE is calculated as the following:

$$MSE_{avg} = \sum_{j=1}^{m} \frac{\sum_{i=1}^{n}(y_i - \hat{y}_i)^2}{m} \qquad (24)$$

Where m is the number of training samples.

Figure 5 demonstrates how the optimization algorithm is applied to help the neural network to update the weights to reach the highest accuracy.

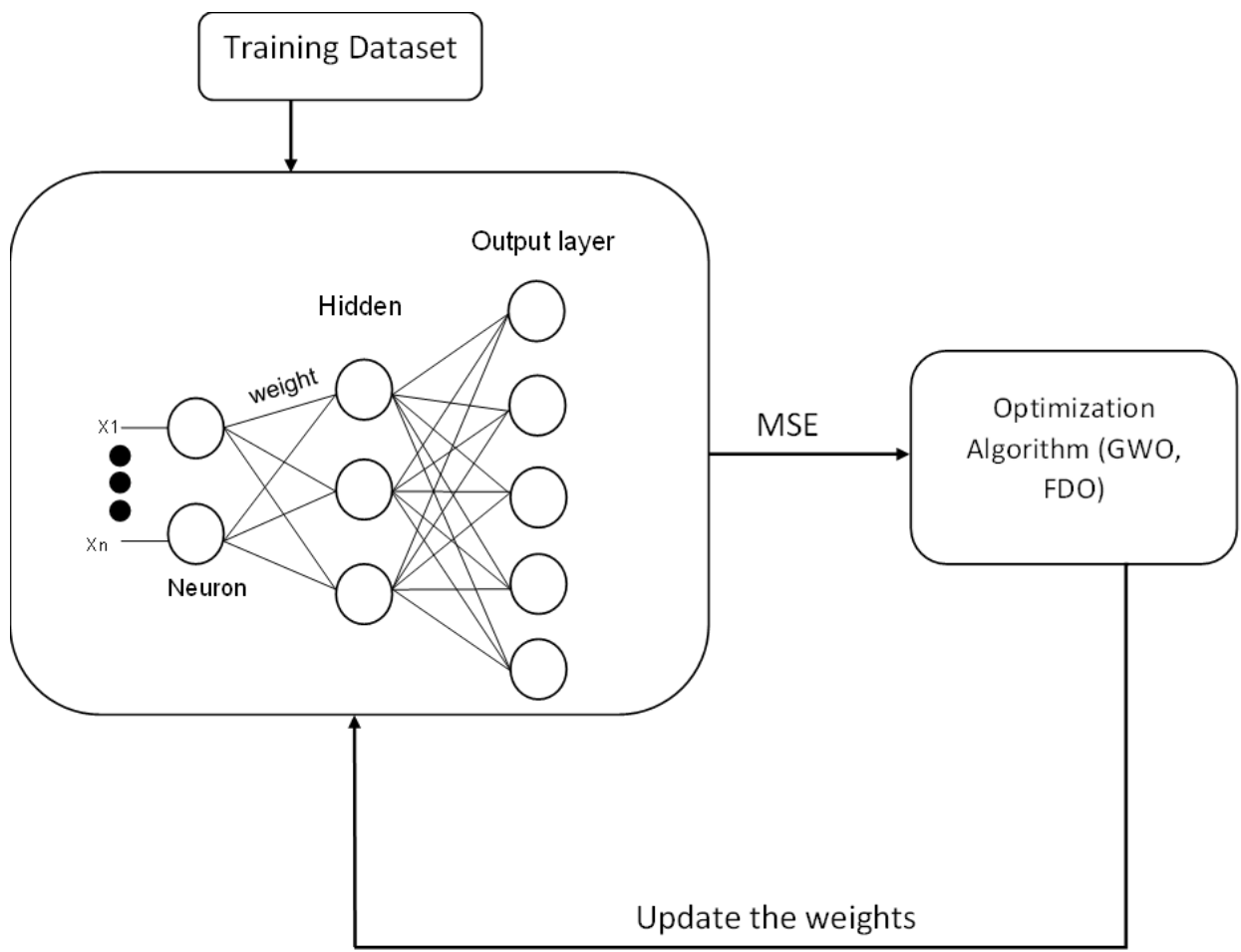

*Figure 4: Training Methods of the proposed models*

## 4.4 Result Evaluation

As mentioned in the previous section. The MSE value is one of the methods that has been used to evaluate the applied models. The models that have very small MSE (close to zero) are considered to be good models.

Another way to evaluate the models is the confusion matrix which is the result of the classification that summarizes the correct and the incorrect predictions. Using the confusion matrix we can get the metrics of sensitivity, specificity, Positive Predictive Value (PPV), and Negative Predictive Value (NPV). These values are calculated to have a better understanding of the model's reliability

[35].

The sensitivity, also known as the true positive rate (TPR), is a metric that measures the proportion of samples that are truly positive and give a positive result. The specificity, also referred to as the true negative rate (TNR), is a metric that measures the proportion of samples that are truly negative and give a negative result. The positive predictive value (PPV) is the probability that a sample that returns a positive result is positive. The negative predictive value (NPV) is the probability that a sample that returns a negative result is negative. The result of these four values is between 0 and 1. the closer value to 1 means the better result it has. In other words, 1 is the best value while 0 is the worst [35].

## 5. Implementation and Result

In this section, we will discuss the achieved results from the implemented architecture and demonstrate the training and testing performance. Section 5.1 describes the experiment environment and the applied framework. Section 5.2 shows the overall performance of the proposed model. The rest of the sections (5.3, 5.4, and 5.5) elaborate on the obtained results of each model in each dataset.

### 5.1. Experimental Setup

We used a Windows system with 16.0 GB of Ram and a 2.00 GHz processor for performing this work. The MATLAB platform was used to build and perform machine learning classification. The classification model was tested on three different datasets. Each dataset was split into an 80:20 ratio. 80% of the data was dedicated to training and the other 20% was dedicated to testing.

### 5.2. Overview

Table 2 shows the result of the correct classification rate for each model in three different datasets. The first dataset contains a total of 3128 samples. 2503 was dedicated to training and the other 625 to testing. The second dataset contained 2102 samples, 1683 dedicated to training and 419 for testing. Lastly, dataset 3, has the highest number of samples with a total of 129,581. 103665 is dedicated to training and 25916 to testing.

Table 2 demonstrated the correct classification rate of training and testing in each model and for each dataset. It also shows the problem dimension (the total number of connections), the search agents that have been used, and the maximum iteration of the search algorithm. In the experiment, we made sure to test all the models under the same conditions and circumstances where we used the same number of iterations and the same number of search agents for all the algorithms in all the models. Ten search agents were used in each algorithm, and the maximum iteration for each was 50 (see Table 3-4). As shown in Table 5, the results obtained by all models are close. However, it can be observed that the models that are trained with the FDO algorithm have a higher chance of getting more accurate results, as they achieved 100% accuracy in all the experiments. However, looking at the run time The FDO takes way longer than the GWO. Observing the run time, we see that the GWO_MLP model takes the shortest time among the other models. The reason for that is in general GWO algorithm is faster than the FDO algorithm and the MLP architecture has fewer connections than the CMLP.

*Table 3: Parameter Setting for GWO*

| Parameter | Value |
| --- | --- |
| No. Search Agents | 10 |
| No. of iterations | 50 |
| Problem dimension | The problem dimension referred to in Table 3 for each model and each dataset |

*Table 4: Parameter Setting for FDO*

| Parameter | Value |
| --- | --- |
| No. Search Agents | 10 |
| No. of iterations | 50 |
| Weight factor | 0 |
| Problem dimension | The problem dimension referred to in Table 3 for each model and each dataset |

*Table 5: Correct classification rate of the proposed models.*

| Model | Dataset | Samples | Dimension | Run time | Training Rate % | Testing Rate % |
|---|---|---|---|---|---|---|
| GWO_MLP | Dataset 1 | 3128 | 253 | 126.96s | 92.8486 | 92.96 |
| GWO_MLP | Dataset 2 | 2102 | 741 | 123.116s | 99.8217 | 99.5227 |
| GWO_MLP | Dataset 3 | 129,81 | 406 | 8641.778s | 99.9923 | 99.9807 |
| MGWO_MLP | Dataset 1 | 3128 | 253 | 112.547s | 95.9249 | 96.48 |
| MGWO_MLP | Dataset 2 | 2102 | 741 | 230.457s | 99.7623 | 99.0453 |
| MGWO_MLP | Dataset 3 | 129,81 | 406 | 9009.343s | 99.9749 | 100 |
| GWO_CMLP | Dataset 1 | 3128 | 264 | 180.037s | 95.3256 | 94.88 |
| GWO_CMLP | Dataset 2 | 2102 | 760 | 173.491s | 99.8812 | 99.7613 |
| GWO_CMLP | Dataset 3 | 129,81 | 420 | 10046.7s | 99.8717 | 100 |
| FDO_MLP | Dataset 1 | 3128 | 253 | 3056.814s | 100 | 100 |
| FDO_MLP | Dataset 2 | 2102 | 741 | 5628.009s | 100 | 100 |
| FDO_MLP | Dataset 3 | 129,81 | 406 | 170856.387 s | 100 | 100 |

## 5.3 Dataset 1 Results

Table 6 shows the performance of the proposed models in terms of Mean Square Error (MSE) and the classification rate of both testing and training. The highest obtained classification rate for training is 100% by FDO_MLP, then FDO_CMLP with 99.88%, followed by MGWO_MLP with 95.92%, GWO_CMLP (95.32%), and lastly, GWO_MLP with the lowest rate of (92.84%). The highest obtained classification rate for testing is 100% by both FDO_MLP and FDO_CMLP, then MGWO_MLP with 97.76%, followed by GWO_CMLP with 97.28%, and lastly, GWO_MLP with the lowest rate of (96.48%). In addition to the classification accuracy Table, 7 shows the evaluation metric of the confusion matrices of the proposed models for dataset 1. Figure 6 shows the roc curve result of all the proposed models tested on dataset 1.

*Table 6: Performance of the proposed models in dataset 1.*

| Model | Training/Testing | Positive case | | | Negative case | | | MSE | Rate % |
|---|---|---|---|---|---|---|---|---|---|
| | | Cases No. | Correct predicts | Accuracy | Cases No. | Correct predicts | Accuracy | | |
| GWO_MLP | Training | 1260 | 1163 | 92.3016 | 1243 | 1161 | 93.4031 | 0.0027385 | 92.8486 |
| | Testing | 304 | 304 | 100 | 321 | 299 | 93.1464 | 0.002745 | 96.48 |
| MGWO_MLP | Training | 1260. | 1230 | 97.619 | 1243 | 1171 | 94.2076 | 0.0018229 | 95.9249 |
| | Testing | 304 | 304 | 100 | 321 | 307 | 95.6386 | 0.001929 | 97.76 |
| GWO_CMLP | Training | 1260. | 1210 | 96.0317 | 1243 | 1176 | 94.6098 | 0.0023577 | 95.3256 |
| | Testing | 304 | 304 | 100 | 321 | 304 | 94.704 | 0.0024762 | 97.28 |
| FDO_MLP | Training | 1260 | 1260 | 100 | 1243 | 1243 | 100 | 0.0022916 | 100 |
| | Testing | 304 | 304 | 100 | 321 | 321 | 100 | 0.0024932 | 100 |
| FDO_CLMP | Training | 1260 | 1257 | 99.7619 | 1243 | 1243 | 100 | 0.0020373 | 99.8801 |
| | Testing | 304 | 304 | 100 | 321 | 321 | 100 | 0.001997 | 100 |

*Table 7: Evaluation of the confusion matrices of the proposed models for dataset 1.*

| Model | Sensitivity | Specificity | PPV | NPV | Accuracy |
|---|---|---|---|---|---|
| GWO_MLP | 1 | 0.93 | 0.93 | 1 | 96.48% |
| MGWO_MLP | 1 | 0.95 | 0.95 | 1 | 97.76% |
| GWO_CMLP | 1 | 0.94 | 0.94 | 1 | 97.28% |
| FDO_MLP | 1 | 1 | 1 | 1 | 100% |
| FDO_CLMP | 1 | 1 | 1 | 1 | 100% |

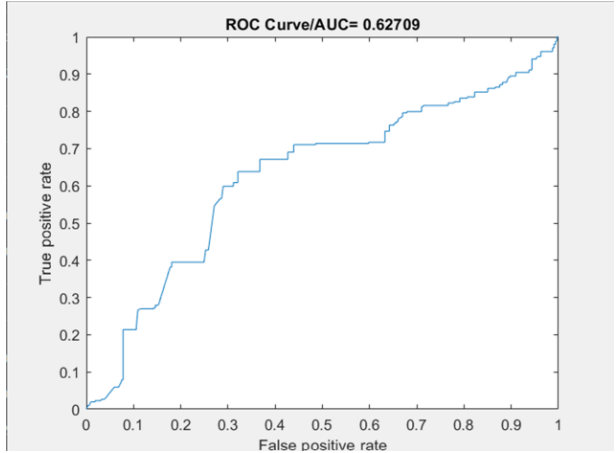

(a) GWO_MLP

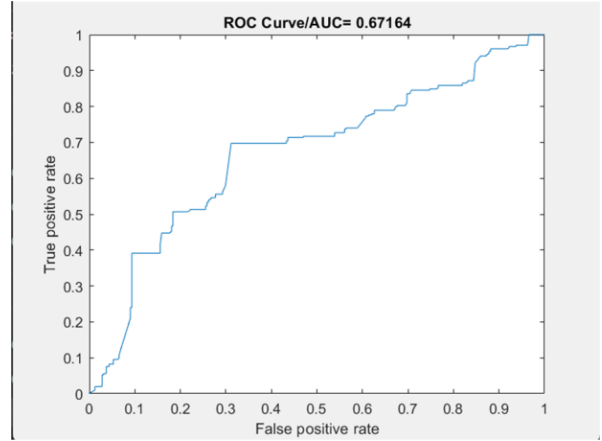

(b) GWO_CMLP

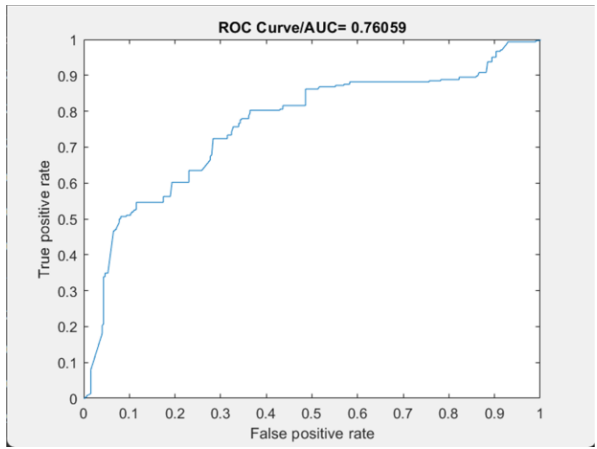

(c) MGWO_MLP

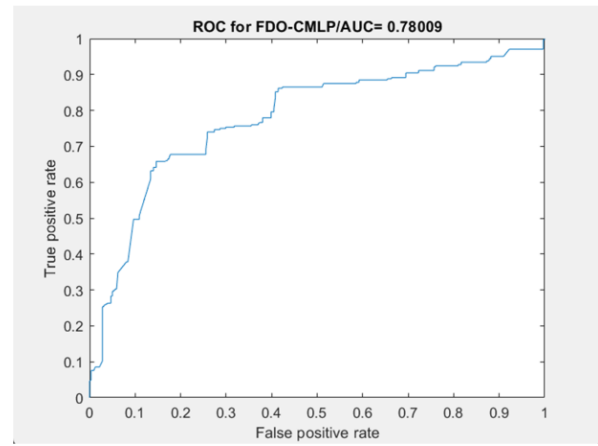

(d) FDO_MLP

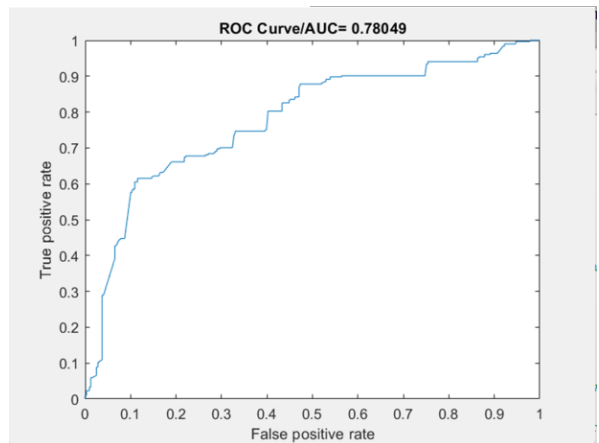

(e) FDO_CMLP

## 5.4 Dataset 2 Results

Table 8 shows the performance of the proposed models in terms of Mean Square Error (MSE) and the classification rate of both testing and training. The highest obtained classification rate for training is 100% by FDO_MLP and FDO_CMLP, followed by GWO_CMLP with 99.88%, then GWO_MLP (99.82%), and lastly, MGWO_MLP with the lowest rate of (99.76%). The highest obtained classification rate for testing is 100% by both FDO_MLP and FDO_CMLP, then GWO_MLP and GWO_CMLP with 99.76%, and finally, MGWO_MLP with the lowest rate of (99.04%). In addition to the classification accuracy, Table 9 shows the evaluation metric of the confusion matrices of the proposed models for dataset 2. Figure 7 shows the roc curve result of all the proposed models tested on dataset 2.

*Table 8: Performance of the proposed models in dataset 2.*

| Model | Training/Testing | Positive case | | | Negative case | | | MSE | Rate |
|---|---|---|---|---|---|---|---|---|---|
| | | Cases No. | Correct predicts | Accuracy | Cases No. | Correct predicts | Accuracy | | |
| GWO_MLP | Training | 842 | 841 | 99.8812 | 841 | 839 | 99.7622 | 0.00057195 | 99.8217 |
| | Testing | 209 | 209 | 100 | 210 | 209 | 99.5238 | 0.00065612 | 99.7613 |
| MGWO_MLP | Training | 842 | 842 | 100 | 841 | 837 | 99.5244 | 0.00071933 | 99.7623 |
| | Testing | 209 | 209 | 100 | 210 | 206 | 98.0952 | 0.00075144 | 99.0453 |
| GWO_CMLP | Training | 842 | 842 | 100 | 841 | 839 | 99.7622 | 0.00068411 | 99.8812 |
| | Testing | 209 | 209 | 100 | 210 | 209 | 99.5238 | 0.00063135 | 99.7613 |
| FDO_MLP | Training | 842 | 842 | 100 | 841 | 841 | 100 | 0.00048108 | 100 |
| | Testing | 209 | 209 | 100 | 210 | 210 | 100 | 0.00085843 | 100 |
| FDO_CLMP | Training | 842 | 842 | 100 | 841 | 841 | 100 | 0.00045953 | 100 |
| | Testing | 209 | 209 | 100 | 210 | 210 | 100 | 0.00048264 | 100 |

*Table 9: Evaluation of the confusion matrices of the proposed models for dataset 2.*

| Model | Sensitivity | Specificity | PPV | NPV | Accuracy |
|---|---|---|---|---|---|
| GWO_MLP | 1 | 1 | 1 | 1 | 99.76% |
| MGWO_MLP | 0.98 | 1 | 1 | 0.98 | 99.05% |

| GWO_CMLP | 0.99 | 1 | 1 | 0.99 | 99.76% |
|---|---|---|---|---|---|
| FDO_MLP | 1 | 1 | 1 | 1 | 100% |
| FDO_CLMP | 1 | 1 | 1 | 1 | 100% |

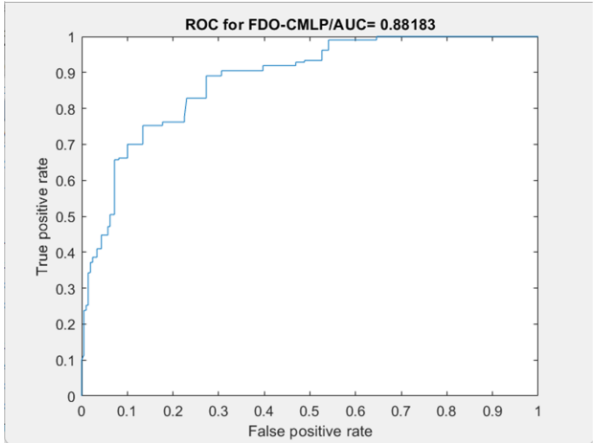

(a) GWO_MLP

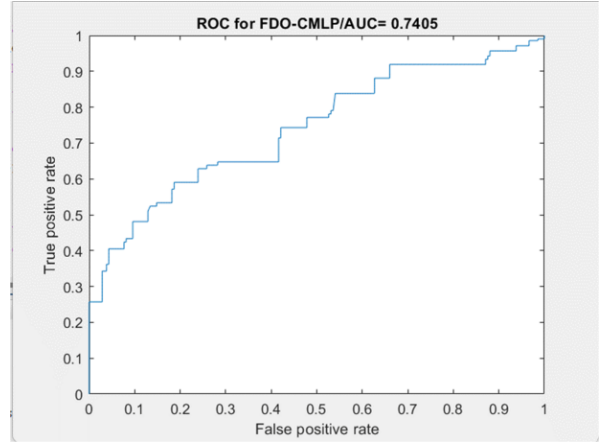

(b) GWO_CMLP

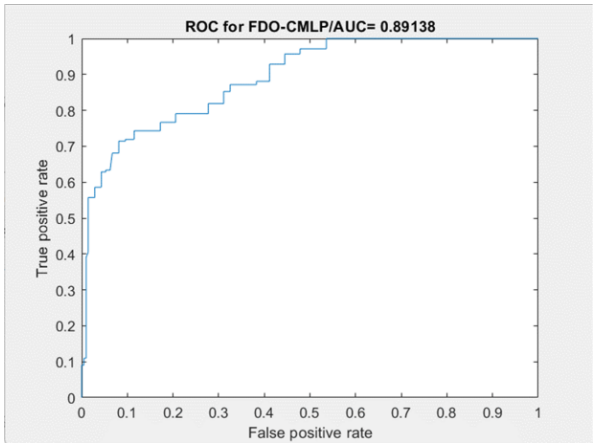

(c) MGWO_MLP

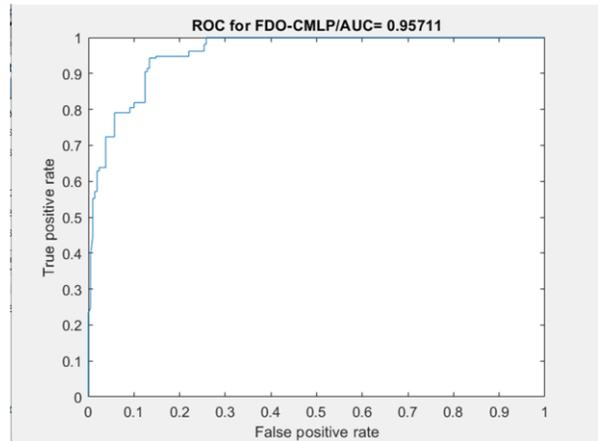

(d) FDO_MLP

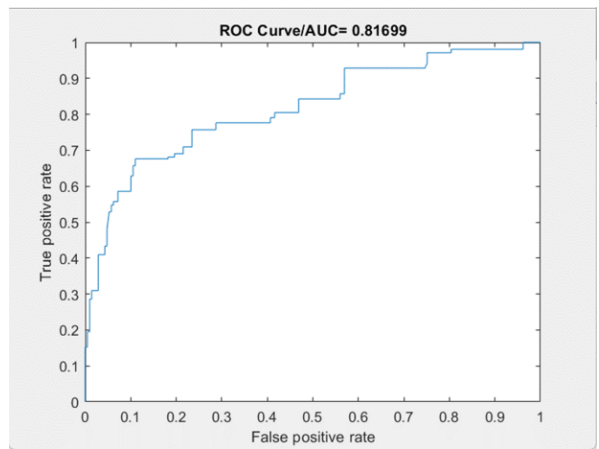

(e) FDO_CMLP

Cite as: Maryam T. Abdulkhaleq, Tarik A. Rashid, Bryar A. Hassan, Abeer Alsadoon, Nebojsa Bacanin, Amit Chhabra, S. Vimal, 2023. Fitness dependent optimizer with neural networks for COVID-19 patients, Computer Methods and Programs in Biomedicine Update, Volume 3, 100090, DOI: https://doi.org/10.1016/j.cmpbup.2022.100090## 5.5 Dataset 3 Results

Table 10 shows the performance of the proposed models in terms of Mean Square Error (MSE) and the classification rate of both testing and training. The performance of classifying this dataset was very promising as this is the largest dataset that has been used in this work. In general, the obtained classification rate is between 99% and 100%. In addition to the classification accuracy, Table 11 shows the evaluation metric of the confusion matrices of the proposed models for dataset 3. Figure 8 shows the roc curve result of all the proposed models tested on dataset 3.

*Table 10: Performance of the proposed models in dataset 3.*

| Model | Training / Testing | Positive case | | | Negative case | | | MSE | rate |
|---|---|---|---|---|---|---|---|---|---|
| | | Cases No. | Correct predicts | Accuracy | Cases No. | Correct predicts | Accuracy | | |
| GWO_MLP | Training | 52396 | 52388 | 99.9847 | 51269 | 51269 | 100 | 0.0015249 | 99.9923 |
| | Testing | 6406 | 6406 | 100 | 19510 | 19507 | 99.9846 | 0.0016695 | 99.9807 |
| MGWO_MLP | Training | 52396 | 52370 | 99.9504 | 51269 | 51269 | 100 | 0.0015553 | 99.9749 |
| | Testing | 6406 | 6406 | 100 | 19510 | 19510 | 100 | 0.0015912 | 100 |
| GWO_CMLP | Training | 52396 | 52263 | 99.7462 | 51269 | 51269 | 100 | 0.001697 | 99.8717 |
| | Testing | 6406 | 6406 | 100 | 19510 | 19510 | 100 | 0.0020708 | 100 |
| FDO_MLP | Training | 52396 | 52396 | 100 | 51269 | 51269 | 100 | 0.0014373 | 100 |
| | Testing | 6406 | 6406 | 100 | 19510 | 19510 | 100 | 0.0015399 | 100 |
| FDO_CLMP | Training | 52396 | 52396 | 100 | 51269 | 51269 | 100 | 0.0015387 | 100 |
| | Testing | 6406 | 6406 | 100 | 19510 | 19510 | 100 | 0.0016973 | 100 |

*Table 11: Evaluation of the confusion matrices of the proposed models for dataset 3.*

| Model | Sensitivity | Specificity | PPV | NPV | Accuracy |
|---|---|---|---|---|---|
| GWO_MLP | 0.99 | 1 | 1 | 0.99 | 99.99% |
| MGWO_MLP | 1 | 1 | 1 | 1 | 100% |
| GWO_CMLP | 1 | 1 | 1 | 1 | 100% |
| FDO_MLP | 1 | 1 | 1 | 1 | 100% |
| FDO_CLMP | 1 | 1 | 1 | 1 | 100% |

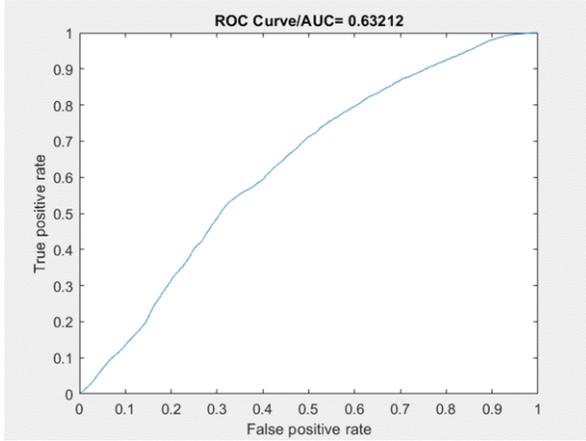

(a) GWO_MLP

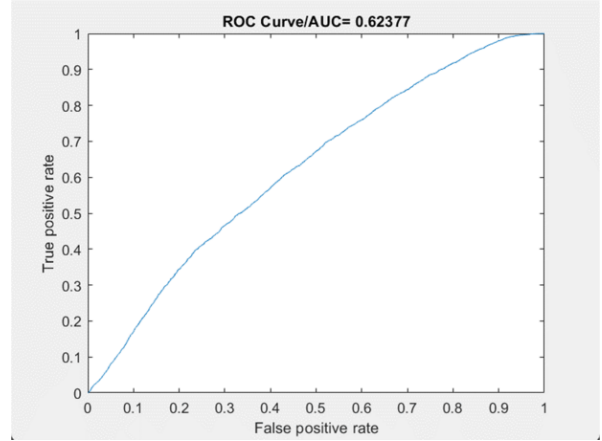

(b) GWO_CMLP

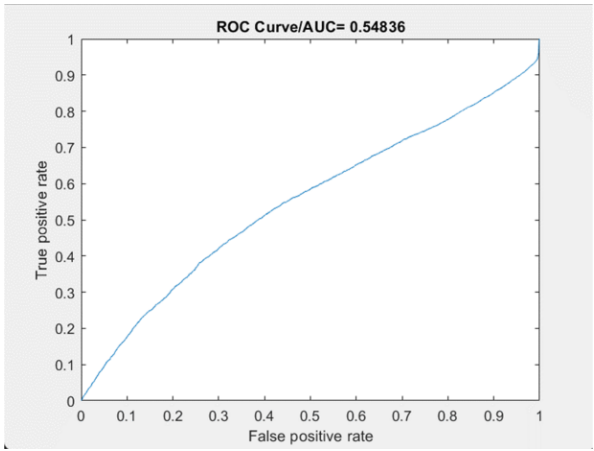

(c) MGWO_MLP

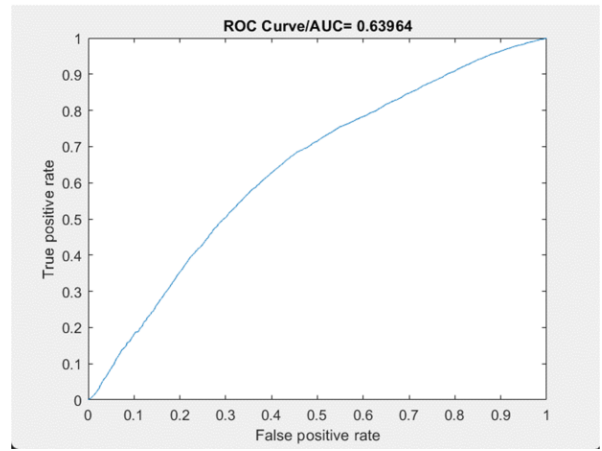

(d) FDO_MLP

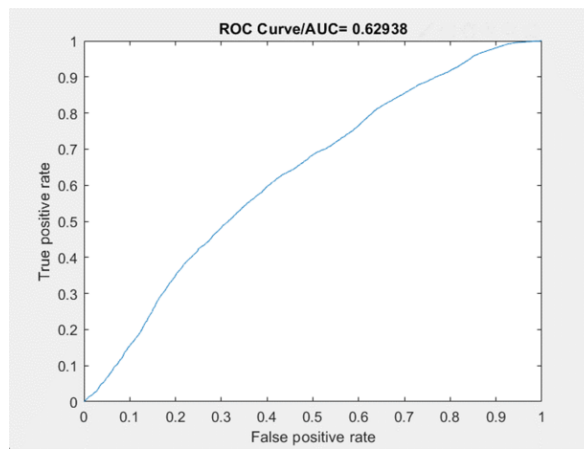

(e) FDO_CMLP

*Figure 8: ROC curve results for dataset 3.*

## 6. Conclusion

Given the rapid spread of COVID-19, which infected millions of people within a few months, the urgent need to make a quick and accurate diagnosis of patients has become urgent. Early diagnosis helps to reduce the spread of the disease and the transmission of infection to other people. ML is one of the non-clinical methods that is considered an alternative means of diagnosing infected patients or predicting disease. The goal of developing such technologies is to reduce the burden that the epidemic has caused on health centers around the world. In this work, we proposed five different ML models for COVID-19 prediction. The purpose of these models is to classify infected and non-infected cases. We relied on two different ANN architectures for analysing COVID-19-related infection data. We also used different training algorithms and different datasets to examine the performance of each model in each type of setting and experiment.

According to the obtained results, the FDO_MLP and FDO_CMLP outperformed the other models in all the performed experiments with an accuracy that reached 100%. However, the FDO algorithm takes more run time than GWO in all three applied datasets. Despite that, in the medical area, the priority is the accuracy of the system, which makes the FDO preferable. The performance of the other models varies slightly according to the type of data applied to them. It is noticed that there is no clear difference between the models that are built based on MLP or CMLP because the obtained results vary from one experiment to another.

The classification models that have been presented in this research show very high precision. As a first step in building a predictive system that can be used in the medical field, the obtained results were promising. The main benefit of developing a classification model is to identify the infected or high-risk patients easily, which is useful in controlling this infection. Despite that, the medical field is very sensitive and requires a highly efficient and reliable system. It is worth noting that getting 100% accuracy is uncommon in classification models. This could be due to several reasons. One of the reasons is the number of features used in the classification and how easy the dataset is to classify. Also, the dataset that has been used doesn't contain noise data, which can lead to producing generalization errors.

Therefore, more tests must be done to ensure the reliability of these models. This research can be expanded or developed by doing the following as future work:

- Getting an extended database that contains more features and more samples. This will make. Considering more features related to COVID-19 will make the prediction process more efficient and comprehensive.

- Building other neural network models that could use the same training algorithms to create a system that can deal with different types of data, such as image, audio, and time-series data. Processing this type of data will provide a more intelligent system that has more benefits, especially in the health sector.

- Exploring new methods will lead to producing more reliable and sophisticated models. For this reason, we need to develop the applied algorithms by modifying them or hybridizing them with other classical or metaheuristic algorithms to produce other models.